\documentclass{article}

\usepackage{iclr2027_conference,times}

\usepackage{amsmath,amsfonts,bm}

\def\eqref#1{equation~\ref{#1}}

\def\1{\bm{1}}

\DeclareMathAlphabet{\mathsfit}{\encodingdefault}{\sfdefault}{m}{sl}
\SetMathAlphabet{\mathsfit}{bold}{\encodingdefault}{\sfdefault}{bx}{n}

\usepackage{amssymb}
\usepackage{array}
\usepackage{booktabs}
\usepackage{longtable}
\usepackage{graphicx}
\usepackage{multirow}
\usepackage{xcolor}
\usepackage{tikz}
\usetikzlibrary{arrows.meta,calc,positioning}
\usepackage{tcolorbox}
\usepackage{hyperref}
\usepackage{url}
\usepackage{xspace}
\hypersetup{hidelinks}

\definecolor{sbblue}{HTML}{DCEBFA}
\definecolor{sborange}{HTML}{FCE4C7}
\definecolor{sbgreen}{HTML}{DDEEDC}
\definecolor{sbpurple}{HTML}{E8DDF4}
\definecolor{sbgray}{HTML}{ECECEC}
\definecolor{sbred}{HTML}{F8D8D8}

\usepackage[ampersand]{easylist}
\ListProperties(Hide2=1,Hide3=2,Progressive*=.5cm,Numbers3=l, Numbers4=r,FinalMark3={)})
\usepackage{etoolbox}
\AtBeginEnvironment{easylist}{\ListProperties(Start1=1)}

\usepackage{xcolor}

\definecolor{cblue}{RGB}{8, 85, 153}

\usepackage{amsmath}
\usepackage[capitalise]{cleveref}
\usepackage{float}
\usepackage{placeins} 
\usepackage{adjustbox}

\usepackage{makecell}
\usepackage[ampersand]{easylist}
\usepackage{todonotes}
\usepackage{caption}
\usepackage{xspace}
\usepackage{markdown}
\usepackage{multirow}
\usepackage{multicol}
\usepackage{graphicx}
\usepackage{subcaption}
\usepackage{titlesec}
\usepackage{tabularx}
\usepackage{array}

\newcommand{\sleuthbench}{\textsc{SleuthBench}\xspace}
\newcommand{\baseD}{\mathcal{D}}
\newcommand{\injD}{\widetilde{\mathcal{D}}}
\newcommand{\effects}{\mathbf{e}}
\newcommand{\correct}{\textsc{Correct}}
\newcommand{\partialgrade}{\textsc{Partial}}
\newcommand{\incorrect}{\textsc{Incorrect}}

\newcommand{\SnapshotResponses}{840}
\newcommand{\SnapshotPythonCorrect}{543}

\newcommand{\SnapshotPythonAccuracy}{64.6}
\newcommand{\SnapshotEmpiricalCorrect}{644}

\newcommand{\SnapshotEmpiricalAccuracy}{76.7}

\newcommand{\SnapshotTotalRecords}{1680}

\newcommand{\SnapshotDatasetPairs}{70}

\newcommand{\SnapshotFramingResponses}{420}
\newcommand{\SnapshotQuestionsPerModel}{140}
\newcommand{\SnapshotEmpiricalUsage}{611}
\newcommand{\SnapshotEmpiricalUsagePercent}{72.7}
\newcommand{\SnapshotEmpiricalCalls}{4120}

\newcommand{\SnapshotBusinessPythonAccuracy}{62.1}

\newcommand{\SnapshotBusinessDelta}{14.8}

\newcommand{\SnapshotDSPythonAccuracy}{67.1}

\newcommand{\SnapshotDSDelta}{9.3}

\newcommand{\SnapshotSonnetPythonAccuracy}{50.7}

\newcommand{\SnapshotOpusDelta}{16.4}

\newcommand{\SnapshotSolPythonAccuracy}{80.0}

\newcommand{\SnapshotSolDelta}{2.1}

\newcommand{\SnapshotDataQualityPythonAccuracy}{83.8}

\newcommand{\SnapshotDataQualityEmpiricalAccuracy}{84.0}

\newcommand{\SnapshotDataQualityGain}{1}

\newcommand{\SnapshotFeatureEffectPythonCorrect}{161}

\newcommand{\SnapshotFeatureEffectPythonAccuracy}{41.9}
\newcommand{\SnapshotFeatureEffectEmpiricalCorrect}{261}

\newcommand{\SnapshotFeatureEffectEmpiricalAccuracy}{68.0}

\newcommand{\SnapshotDataQualityQuestionsPerModel}{76}
\newcommand{\SnapshotFeatureEffectQuestionsPerModel}{64}

\newcommand{\SnapshotFeatureEffectBusinessPythonAccuracy}{32.8}
\newcommand{\SnapshotFeatureEffectBusinessEmpiricalAccuracy}{65.1}
\newcommand{\SnapshotFeatureEffectBusinessDelta}{32.3}
\newcommand{\SnapshotFeatureEffectDSPythonAccuracy}{51.0}
\newcommand{\SnapshotFeatureEffectDSEmpiricalAccuracy}{70.8}
\newcommand{\SnapshotFeatureEffectDSDelta}{19.8}

\newcommand{\SnapshotSonnetDataQualityPythonAccuracy}{77.6}

\newcommand{\SnapshotSonnetFeatureEffectPythonAccuracy}{18.8}

\newcommand{\SnapshotOpusFeatureEffectDelta}{35.9}

\newcommand{\SnapshotFableDataQualityDelta}{2.6}

\newcommand{\SnapshotSolDataQualityPythonAccuracy}{86.8}

\newcommand{\SnapshotSolFeatureEffectPythonAccuracy}{71.9}

\newcommand{\SnapshotSolFeatureEffectEmpiricalAccuracy}{78.1}
\newcommand{\SnapshotSolFeatureEffectDelta}{6.3}

\newcommand{\SnapshotBaselineRows}{%
Claude Sonnet 5 & 77.6 & 18.8 & 50.7 \\
Claude Opus 4.8 & 84.2 & 29.7 & 59.3 \\
Claude Fable 5 & 85.5 & 45.3 & 67.1 \\
GPT-5.6 Sol & 86.8 & 71.9 & 80.0 \\
GPT-5.6 Terra & 85.5 & 51.6 & 70.0 \\
GPT-5.6 Luna & 82.9 & 34.4 & 60.7 \\
}
\newcommand{\SnapshotBaselineAllRow}{All models & 83.8 & 41.9 & 64.6 \\}
\newcommand{\SnapshotModelRows}{%
Claude Sonnet 5 & 77.6 & 77.6 & 0.0 & 18.8 & 53.1 & +34.4 & 50.7 & 66.4 & +15.7 \\
Claude Opus 4.8 & 84.2 & 84.2 & 0.0 & 29.7 & 65.6 & +35.9 & 59.3 & 75.7 & +16.4 \\
Claude Fable 5 & 85.5 & 88.2 & +2.6 & 45.3 & 75.0 & +29.7 & 67.1 & 82.1 & +15.0 \\
GPT-5.6 Sol & 86.8 & 85.5 & -1.3 & 71.9 & 78.1 & +6.3 & 80.0 & 82.1 & +2.1 \\
GPT-5.6 Terra & 85.5 & 85.5 & 0.0 & 51.6 & 82.8 & +31.3 & 70.0 & 84.3 & +14.3 \\
GPT-5.6 Luna & 82.9 & 82.9 & 0.0 & 34.4 & 53.1 & +18.8 & 60.7 & 69.3 & +8.6 \\
}
\newcommand{\SnapshotOverallRows}{%
All models & 83.8 & 84.0 & +0.2 & 41.9 & 68.0 & +26.0 & 64.6 & 76.7 & +12.0 \\
}
\newcommand{\SnapshotFramingRows}{%
Business & 62.1 & 76.9 & +14.8 \\
Data science & 67.1 & 76.4 & +9.3 \\
}

\newcommand{\SnapshotFamilyGradeRows}{%
Data quality & 456 & 382/24/50 & 383/12/61 \\
Feature contribution & 384 & 161/31/192 & 261/39/84 \\
Total & 840 & 543/55/242 & 644/51/145 \\
}
\newcommand{\SnapshotFamilyFramingRows}{%
Data quality & Business & 228 & 86.8 & 86.8 & 0.0 \\
Data quality & Data science & 228 & 80.7 & 81.1 & +0.4 \\
Feature contribution & Business & 192 & 32.8 & 65.1 & +32.3 \\
Feature contribution & Data science & 192 & 51.0 & 70.8 & +19.8 \\
}

\newcommand{\SnapshotModelFramingRows}{%
Claude Sonnet 5 & Business & 70 & 47.1 & 67.1 & +20.0 \\
Claude Sonnet 5 & Data science & 70 & 54.3 & 65.7 & +11.4 \\
Claude Opus 4.8 & Business & 70 & 52.9 & 74.3 & +21.4 \\
Claude Opus 4.8 & Data science & 70 & 65.7 & 77.1 & +11.4 \\
Claude Fable 5 & Business & 70 & 62.9 & 84.3 & +21.4 \\
Claude Fable 5 & Data science & 70 & 71.4 & 80.0 & +8.6 \\
GPT-5.6 Sol & Business & 70 & 87.1 & 81.4 & -5.7 \\
GPT-5.6 Sol & Data science & 70 & 72.9 & 82.9 & +10.0 \\
GPT-5.6 Terra & Business & 70 & 68.6 & 87.1 & +18.6 \\
GPT-5.6 Terra & Data science & 70 & 71.4 & 81.4 & +10.0 \\
GPT-5.6 Luna & Business & 70 & 54.3 & 67.1 & +12.9 \\
GPT-5.6 Luna & Data science & 70 & 67.1 & 71.4 & +4.3 \\
}
\newcommand{\SnapshotGradeRows}{%
Claude Sonnet 5 & 140 & 71/12/57 & 93/8/39 \\
Claude Opus 4.8 & 140 & 83/9/48 & 106/8/26 \\
Claude Fable 5 & 140 & 94/8/38 & 115/10/15 \\
GPT-5.6 Sol & 140 & 112/9/19 & 115/9/16 \\
GPT-5.6 Terra & 140 & 98/9/33 & 118/6/16 \\
GPT-5.6 Luna & 140 & 85/8/47 & 97/10/33 \\
}

\title{\sleuthbench: Benchmarking Statistical LLM Evaluation Using Tabular Hidden Signals}

\author{
Jingyun Jia \\
University of Wisconsin--Madison, Intelligible \\
\texttt{jjia39@wisc.edu}
\And
Antoine Remond-Tiedrez \\
Intelligible \\
\texttt{antoine@intelligible.ai}
\AND
Aaron Alvarez \\
University of Cincinnati \\
\texttt{aaronalvarez.swe@gmail.com}
\And
Joshua Shunk \\
Intelligible \\
\texttt{jshunk@stanford.edu}
\AND
Rich Caruana \\
Intelligible \\
\texttt{richcaruana@gmail.com}
\And
Ben Lengerich \\
University of Wisconsin--Madison, Intelligible \\
\texttt{lengerich@wisc.edu}
}

\iclrfinalcopy

\begin{document}

\maketitle

\lhead{Preprint}

\begin{abstract}
Evaluating statistical discovery by large language model (LLM) agents
requires verifiable analytical ground truth. Establishing such ground truth
for real-world datasets is costly, and prior knowledge of public datasets
can influence agent responses. We introduce \sleuthbench, a benchmark that
addresses both problems by injecting controlled data-quality problems and
feature effects into public tabular datasets: the injected pattern
determines the answer, so reference answers are computed automatically and
memorized knowledge of the original table is insufficient, while the table
keeps its background structure. The injected patterns are modeled on
phenomena reported in real data analyses. The benchmark defines 17 question
templates in two families: data-quality questions and feature-contribution
questions. We evaluate six state-of-the-art LLMs that analyze the data using a Python
coding tool, on data-science and business phrasings of
\SnapshotDatasetPairs{} validated dataset--template combinations, yielding \SnapshotTotalRecords{} graded responses in total. The models
detect data-quality problems reliably (\SnapshotDataQualityPythonAccuracy\%
accuracy) but recover feature contributions poorly
(\SnapshotFeatureEffectPythonAccuracy\%). Finding how features shape the
target requires searching over both candidate variables and analytical
procedures. To
address this issue, we propose the {\it Empirical Layer}, a set of precomputed
statistical artifacts comprising summaries, fitted feature and interaction
effects, and dataset descriptions, which exposes candidate patterns for
direct inspection. Access to these artifacts raises feature-contribution accuracy from \SnapshotFeatureEffectPythonAccuracy\%
to \SnapshotFeatureEffectEmpiricalAccuracy\%.
\end{abstract}

\section{Introduction}
\label{sec:introduction}

Large language model (LLM) agents increasingly perform data analysis by inspecting tables, writing
code, identifying statistical relationships, and explaining their findings.
Evaluating these agents requires determining whether their answers are
supported by the supplied data. Public datasets complicate this assessment:
prior work shows that LLMs can memorize popular tabular datasets and perform
better on datasets encountered during pretraining
\citep{bordt2024elephants}. Consequently, accuracy on a static public dataset
can reflect both prior dataset knowledge and the ability to recover and
interpret evidence from the table.

Early table-reasoning benchmarks primarily evaluate retrieval and operations over table values, such as selecting rows, comparing or aggregating entries, and verifying whether a statement is supported by a table \citep{pasupat2015compositional,chen2020tabfact}. More recent benchmarks broaden this scope to include statistical and data-analysis questions \citep{wu2024tablebench}. However, they do not systematically test whether an agent can recover a controlled statistical phenomenon distributed across observations. \sleuthbench targets this distinction by evaluating patterns—such as nonlinear effects, feature interactions, and data-quality artifacts—that are implicit in the empirical distribution rather than stated in any individual cell.

Establishing reliable reference answers for statistical questions over real tables can also require substantial expert effort, making such benchmarks costly to build and refresh. 
\sleuthbench addresses this challenge by introducing a prescribed
statistical phenomenon into a public table and generating a corresponding
question. The intervention can introduce a single-feature effect, a pairwise
interaction, or a data-quality anomaly while retaining the surrounding
table structure. A task-specific validator checks that the intended
phenomenon is detectable in the modified table and that competing
patterns do not make the answer ambiguous.

\sleuthbench addresses public-data memorization and costly ground-truth curation by injecting controlled statistical phenomena into source tables and automatically computing reference answers for validated instances. These interventions make knowledge of the original table insufficient to determine the answer, while automatic answer computations avoid the need to manually label each instance. The source supplies correlations, mixed data types, and competing patterns; the intervention specifies the target phenomenon and its acceptance criterion. This creates a middle ground between manually labeling observational data and synthesizing an entire table: the benchmark retains a source table's background structure while controlling the answer-defining evidence. Regenerating interventions with new random seeds reduces dependence on a fixed set of public answers that future LLM could train on.

These choices allow us to evaluate whether agents can identify
statistical patterns across observations while keeping reference
answers verifiable and evaluation instances refreshable.
Our contributions are:

\begin{itemize}
    \item We introduce a benchmark construction framework that injects statistical
    phenomena into public tables, validates their observable effects, and
    computes reference answers independently of the evaluated LLM and its
    analysis tools. Because the injected phenomenon determines the answer,
    memorized knowledge of the original table is insufficient, while reusable
    validators and answer computations reduce per-instance curation and support
    refreshed cases.
    \item We provide 17 question templates backed by 14 injectors,
    covering data-quality problems and feature contributions (single-feature
    effects and pairwise interactions).
    Each template is a reusable task specification that defines compatible
    variables, an injection, alternate question framings, an answer format, and
    validation criteria.
    \item We evaluate six LLMs using Python alone and Python augmented
    with the Empirical Layer, which consists of structured statistical evidence about the data.
    This paired comparison tests whether making candidate
    evidence easier to inspect changes statistical-discovery performance,
    informing how analysis interfaces should present candidate patterns.
    With Python alone, the models handle data-quality questions well but
    recover feature contributions poorly; the Empirical Layer raises feature-contribution
    accuracy from \SnapshotFeatureEffectPythonAccuracy\% to
    \SnapshotFeatureEffectEmpiricalAccuracy\% and overall accuracy from
    \SnapshotPythonAccuracy\% to \SnapshotEmpiricalAccuracy\%.
\end{itemize}

\section{Related Work}
\label{sec:related}

Our work has two components: constructing a benchmark through controlled
statistical injections into public source tables, and evaluating how access
to structured statistical evidence affects data analysis agents.
We review related agent systems and benchmarks, followed by exploratory
data analysis and interpretable models that inform the artifacts in our Empirical Layer.

\paragraph{Data analysis agents.}
LLM-based data analysis agents combine natural-language interaction with
planning, code execution, and iterative feedback to carry out analytical
workflows. TaskWeaver supports stateful execution and integrates generated
code with domain-specific plugins \citep{qiao2023taskweaver}. Data Interpreter
represents task dependencies through hierarchical graphs and progressively
verifies and refines intermediate steps \citep{hong2025datainterpreter}.
LAMBDA separates code generation and debugging into programmer and inspector
agents, while supporting user intervention and external analytical methods
\citep{sun2025lambda}. \sleuthbench provides a complementary evaluation
setting for such agents: it specifies the target phenomenon and reference
answer while allowing the agent to choose its analysis procedure.

\paragraph{Data science benchmarks.}
Existing benchmarks assess different stages of analytical work.
WikiTableQuestions and TabFact evaluate compositional question answering and
fact verification over tables \citep{pasupat2015compositional,chen2020tabfact},
while DS-1000 evaluates data science code using functional tests and
implementation constraints \citep{lai2023ds1000}. InfiAgent-DABench and
DA-Code evaluate analysis through interaction with execution environments
\citep{hu2024infiagent,huang2024dacode}; DSBench extends coverage to realistic
analysis and modeling workflows \citep{jing2024dsbench}. QRData focuses on
statistical and causal reasoning \citep{liu2024qrdata}, and DiscoveryBench
evaluates data-driven discovery using both research-derived and synthetic
tasks \citep{majumder2024discoverybench}. DataSciBench addresses complex
evaluation criteria through semi-automated ground-truth construction with
human verification and programmatic evaluation \citep{zhang2026datascibench}.
Benchmark construction also addresses memorization: DS-1000 perturbs source
problems, while LiveBench refreshes questions from recently released sources
\citep{lai2023ds1000,white2025livebench}. \sleuthbench complements these benchmarks through its construction
approach: it injects controlled statistical phenomena into public source
tables, validates the resulting evidence, and computes reference answers deterministically.

\paragraph{Exploratory data analysis and interpretable models.}
Exploratory data analysis uses statistical summaries and graphical
representations to examine distributions, relationships, and unusual
observations, guiding subsequent analysis \citep{tukey1977exploratory}.
Visualization recommendation systems assist the search for informative views.
Voyager 2 combines manual and automatic chart specification to support both
broad exploration and focused questions \citep{wongsuphasawat2017voyager2}.
Foresight ranks candidate insights using statistical properties such as
correlation, skewness, and outliers \citep{demiralp2017foresight}.
Interpretable models provide a complementary approach: generalized additive
models with pairwise interactions, including Explainable Boosting Machines,
decompose predictions into inspectable feature effects and interaction
surfaces \citep{lou2013accurate,nori2019interpretml}. Such models have also
been used to investigate missingness and imputation behavior
\citep{chen2023missing}. We combine statistical summaries and
interpretable-model outputs into the Empirical Layer (defined precisely in Section~\ref{sec:results}), and evaluate how access to these artifacts affects
agents' ability to recover the prescribed phenomena.

\section{\sleuthbench: A Benchmark for Statistical Discovery}
\label{sec:method}

\subsection{Benchmark Overview}

\sleuthbench evaluates whether an agent can recover a prescribed statistical
phenomenon from a supplied table. Each benchmark instance contains three
pieces: an edited table, a natural-language question, and a verifiable
reference answer. We construct an instance by combining a public base table
with a compatible \emph{question template}. A template is a reusable task
specification: it states what statistical pattern to create, which tables and
variables are eligible, what question to ask, and what form the answer should
take. For example, the inverted-U template asks which numerical feature has a
relationship with the target that peaks at an interior value, and where that
peak occurs.

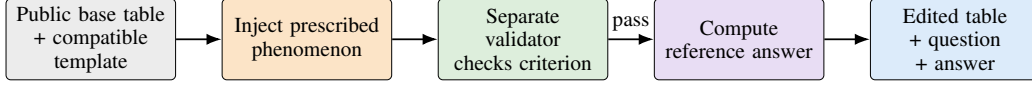
\begin{figure}[H]
\centering
\resizebox{0.98\linewidth}{!}{%
\begin{tikzpicture}[
  font=\small, >=Latex,
  box/.style={draw, rounded corners=2pt, align=center, minimum height=12mm,
    text width=23mm, inner sep=3pt},
  flow/.style={->, line width=0.75pt}
]
\node[box,fill=sbgray] (base) at (0,0) {Public base table\\+ compatible template};
\node[box,fill=sborange] (inject) at (3.2,0) {Inject prescribed\\phenomenon};
\node[box,fill=sbgreen] (validate) at (6.4,0) {Separate validator checks criterion};
\node[box,fill=sbpurple] (compute) at (9.6,0) {Compute reference answer};
\node[box,fill=sbblue] (answer) at (12.8,0) {Edited table \\ + question + answer};
\draw[flow] (base) -- (inject);
\draw[flow] (inject) -- (validate);
\draw[flow] (validate) -- node[above] {pass} (compute);
\draw[flow] (compute) -- (answer);
\end{tikzpicture}}
\caption{\sleuthbench construction. A separate validator
remeasures the injected pattern before a deterministic function computes the
reference answer.}
\label{fig:pipeline}
\end{figure}

Construction proceeds in three steps (Figure~\ref{fig:pipeline}). First, a
randomized injector edits the base table to create the requested pattern and
records what it changed. Second, a separate task-specific validator measures
the edited table and checks that the pattern satisfies the template's
acceptance criteria. Third, a deterministic function computes the
reference answer from the accepted table and intervention record. Only candidates that pass validation and receive a computed reference answer enter the benchmark;
Section~\ref{sec:construction} gives the technical construction, and the
supplementary material describe every template, injection, and question.

\paragraph{Data collection.}
We assemble public tables from the UCI repository, scikit-learn, and Kaggle,
covering maintenance, transportation, housing, and energy use. Our experiments
use 1,000-row tables from six base datasets:
\href{https://archive.ics.uci.edu/dataset/601/ai4i+2020+predictive+maintenance+dataset}{AI4I 2020 Predictive Maintenance},
\href{https://archive.ics.uci.edu/dataset/275/bike+sharing+dataset}{Bike Sharing},
\href{https://scikit-learn.org/stable/modules/generated/sklearn.datasets.fetch_california_housing.html}{California Housing},
\href{https://www.kaggle.com/datasets/harlfoxem/housesalesprediction}{King County Housing},
\href{https://archive.ics.uci.edu/dataset/492/metro+interstate+traffic+volume}{Metro Interstate Traffic Volume},
and \href{https://archive.ics.uci.edu/dataset/851/steel+industry+energy+consumption}{Steel Industry Energy Consumption}.
Each source snapshot is standardized with a designated target column, row identifiers, and a schema summary, and its rows are shuffled. Template compatibility determines which injections can be attempted, while validation determines which resulting instances enter the benchmark. Full integrity checks, preprocessing steps, and additional prepared table sizes are described in Appendix~\ref{app:datasets}.

\paragraph{Phenomena and question framings.}
The benchmark defines 17 templates implemented by 14 injectors in
two template families (Table~\ref{tab:taxonomy}): data-quality questions ask what is
wrong with a table, and feature-contribution questions ask how features shape the
target. Each template specifies compatible
feature types, applicability constraints, a statistical task, and an answer
format. 

Every accepted instance supports two question framings: a data-science
prompt using analytical terminology and a business prompt describing an
operational concern. The framings differ in terminology and in how explicitly
they specify the intended analysis. Both use the same table, selected
variables, and reference answer, enabling paired comparisons of how question
framing affects evidence recovery and interpretation.
Appendix~\ref{app:phenomena} lists the templates and injections.

\begin{table}[H]
\caption{The 17 benchmark question templates. Appendix~\ref{app:phenomena}
(Table~\ref{tab:full-inventory}) lists every template with its injection and
both question wordings.}
\label{tab:taxonomy}
\centering
\small
\setlength{\tabcolsep}{4pt}
\begin{tabular}{>{\raggedright\arraybackslash}p{0.14\linewidth}c
                >{\raggedright\arraybackslash}p{0.38\linewidth}
                >{\raggedright\arraybackslash}p{0.34\linewidth}}
\toprule
Family & Templates & Representative injections & Representative tasks \\
\midrule
Data quality & 8 & Corrupt a subset of target values and add a binary column
marking those rows; replace some category values with a missing-like label;
add a group column with one very rare group. & Identify the marker column or
the missing-like label; decide whether there is sufficient statistical evidence to make claims about the rare group. \\
Feature contribution & 9 & Permute one numerical feature to break its link to the
target; add a smooth inverted-U effect of one
feature on the target; add an XOR-shaped effect on the target over two numerical features. & Name the
noise feature; name the feature and the value at which its effect peaks; name the interacting pair or its conditional direction. \\
\bottomrule
\end{tabular}
\end{table}


\begin{samepage}
\subsection{Phenomenon Injection and Quality Control}
\label{sec:construction}

\paragraph{Constrained intervention.}
Let $\baseD$ denote a public base table, $\tau$ a compatible template, and
$\theta$ the template's intervention parameters. The randomized
injection operator returns
\begin{equation}
  (\injD,\effects)=I_\tau(\baseD;\theta),
\end{equation}
where $\injD$ is the edited table and $\effects$ records the selected variables
and the exact edit. Before applying the operator, template matching checks feature
types, cardinalities, target type, and other task-specific requirements. 
\par
\end{samepage}

\begin{tcolorbox}[colback=sborange, colframe=sborange!30!black, title=Running example: injecting an inverted-U effect]
The inverted-U template applies to a table with a numerical feature
and target. Its injector adds a target effect centered at an interior
feature value and decreasing on both sides.
\end{tcolorbox}

\paragraph{Validation and quality control.}
An edit can execute successfully yet remain ambiguous or be overwhelmed by
the base table. Each candidate therefore undergoes a phenomenon-specific
validation gate $V_\tau(\injD,\effects)$. The validator measures the resulting pattern directly from the final table and accepts a candidate only when the answer-defining pattern
meets the template's operational criterion and, where required, exceeds
competing features or pairs. More details are in Appendix~\ref{app:validation}.

\begin{tcolorbox}[colback=sbgreen, colframe=sbgreen!30!black, title=Running example: validating an inverted-U effect]
After the inverted-U template has been applied, the validator measures the relationship between the selected feature and the target on the final table. It groups the rows into five equal-count bins by feature value and checks three things: the bin with the highest mean target is unique; the injected center falls inside that bin; and the feature shows a clear rise-then-fall pattern.
\end{tcolorbox}

\paragraph{Independent ground truth.}
For an accepted candidate, a template-specific function computes the ground truth
$a^\star=A_\tau(\injD,\effects)$. The template determines whether this answer
is, for example, a feature name, feature pair, row set, or feature--value pair.
Validation determines whether the edited table satisfies the template's
acceptance criteria, whereas answer computation supplies the value used for
grading.

\begin{tcolorbox}[colback=sbpurple, colframe=sbpurple!30!black, title=Running example: computing ground truth for an inverted-U effect]
Once the inverted-U injection has passed the validator, the automatic answer computation provides the reference answer in two parts. The feature name is read directly from the edit record. The peak value is computed on the edited table: the function fits two smooth curves to the feature--target relationship, finds the peak on each curve and averages the two locations. The LLM is then asked the following data science question:

``\emph{One feature has an inverted-U relationship with \texttt{\{OUTCOME\_COL\}} --- performance peaks at a specific value and declines on both sides. Which feature, and at what value is the peak? Return your answer as: column\_name, value}.''
\end{tcolorbox}

\subsection{Injection Design}

The design of the injections began with a simple question: what does a data analyst do when examining a new table? An analyst typically first inspects the data itself, checking for outliers, missing values, and other data-quality problems, and then fits a model to examine how features affect the target. These two stages define the tasks an agent should be able to carry out, and they give the benchmark its two families: data-quality questions, which ask what is wrong with a table, and feature-contribution questions, which ask how features shape the target. Each injection recreates the situation that one of these analyses is meant to detect.

A second consideration is memorization. An analyst who receives a table they have analyzed before, or one that is widely known, may skip parts of the analysis and rely on what they remember about the data; LLMs show the same tendency on popular public datasets \citep{bordt2024elephants}. We therefore inject specific patterns into the features and outcome of each table, so that recovering the injected pattern requires performing the analysis.

In addition, some of our templates draw on findings reported in real data science analyses. For example, in pneumonia data, asthma was associated with lower observed mortality, a pattern that \citet{caruana2015intelligible} attribute to earlier ICU admission and more aggressive treatment. Our categorical-target-outlier injection plants an analogous pattern for a numerical target: one category has an unusually high or low mean outcome. Injections like this plant controlled versions of phenomena observed in real analyses.

\subsection{Comparison with Existing Benchmarks}

Existing benchmarks enable systematic evaluation of agents on
data-analysis workflows and statistical reasoning tasks. However,
challenges remain in balancing ground-truth curation costs with
data realism, mitigating reliance on memorized public data, and
refreshing evaluation instances over time. We discuss how
\sleuthbench addresses these challenges below.

\begin{enumerate}
    \item \textbf{Ground-truth cost and realism.}
    Establishing and verifying reliable reference answers for rich
    analytical tasks over observational data can require substantial
    expert effort
    \citep{liu2024qrdata,majumder2024discoverybench,zhang2026datascibench}.
    Fully synthetic data make labels inexpensive once a generator is
    specified, but their realism depends on that generator.
    \sleuthbench combines public source tables with controlled statistical
    injections. Once the injection, validation, and answer-computation
    procedures are defined, they can be reused across compatible tables
    and seeds, reducing per-instance curation effort while retaining much
    of the source table's background structure.

    \item \textbf{Memorization of source data.}
    Public tables and familiar schemas can be memorized
    \citep{bordt2024elephants}. By changing the answer-defining patterns,
    \sleuthbench makes recall of the original table insufficient to
    determine the answer, although schema knowledge and domain priors
    may still help.

    \item \textbf{Refreshing benchmark instances.}
    Any published benchmark instance can enter later training corpora.
    LiveBench and LiveCodeBench address this by periodically writing new
    questions from newly released material, such as recent papers, news
    articles, datasets, and programming-contest problems, so that the
    questions postdate the models' training cutoffs
    \citep{white2025livebench,jain2024livecodebench}. \sleuthbench can instead simply rerun the pipeline with a new seed and produce fresh instances from the same source tables.

\end{enumerate}

\section{\sleuthbench Remains Challenging for Pretrained LLMs}
\label{sec:experiments}

\subsection{Experimental Setup}

We first test whether pretrained LLMs can recover the prescribed phenomena
using executable analysis alone. Each agent receives the edited table and
question and has only a Python tool. This tool executes agent-generated
Python code on a persistent pandas dataframe containing the edited table,
allowing the agent to inspect the data, perform analyses, and revise its
approach using execution outputs and errors. We provide a Python tool
rather than placing the table directly in the prompt because LLMs alone are unreliable even at basic structural operations on tables such as cell lookup and row retrieval
\citep{sui2024tablemeets,liu2024rethinking}.

We use the six 1,000-row datasets described in
Section~\ref{sec:method}. All six LLM models are evaluated on the same
\SnapshotDatasetPairs{} validated dataset--template combinations, each with
a data science and a business question. Table~\ref{tab:counts} lists the question counts by template family and by framing. All counts and
accuracies in this section and the following section refer to this question set.
Candidates rejected by the validator during the pipeline construction are
excluded from the model evaluation.

\begin{table}[t]
\caption{Questions per model, by template family and
question framing.}
\label{tab:counts}
\centering
\small
\begin{tabular}{lrrr}
\toprule
Framing & Data quality & Feature contribution & Total \\
\midrule
Data science & 38 & 32 & \SnapshotDatasetPairs{} \\
Business & 38 & 32 & \SnapshotDatasetPairs{} \\
\midrule
Total & \SnapshotDataQualityQuestionsPerModel{} & \SnapshotFeatureEffectQuestionsPerModel{} & \SnapshotQuestionsPerModel{} \\
\bottomrule
\end{tabular}
\end{table}

\paragraph{Scoring.}
Responses are graded \correct, \partialgrade, or \incorrect{} against the
reference answer using a template-specific rubric. Numerical and
column--value answers use a 5\% relative tolerance; structured answers are more exact and must
identify the requested columns, category, or row set. The pipeline
combines direct checks with LLM-based assessment of answers requiring
interpretation \citep{zheng2023judge}. Reported accuracy counts only
\correct{} responses and is a percentage.

\paragraph{Models.}

We evaluate six frontier LLMs, Claude Sonnet 5, Claude Opus 4.8, Claude Fable 5, GPT-5.6 Sol,
GPT-5.6 Terra, and GPT-5.6 Luna under the setup above.

\subsection{Results and Challenges}

Python access alone leaves substantial errors across the six models (Table~\ref{tab:baseline}). Accuracy pooled across the data quality and feature contribution injection families is only
\SnapshotPythonAccuracy\% (\SnapshotPythonCorrect{} correct answers),
with individual scores ranging from \SnapshotSonnetPythonAccuracy\% to
\SnapshotSolPythonAccuracy\%. On data-quality questions, every model performs
well: pooled accuracy is \SnapshotDataQualityPythonAccuracy\%, and per-model
accuracy ranges from \SnapshotSonnetDataQualityPythonAccuracy\% to
\SnapshotSolDataQualityPythonAccuracy\%. On feature-contribution questions, however, pooled
accuracy drops to \SnapshotFeatureEffectPythonAccuracy\%, with per-model
accuracy ranging from \SnapshotSonnetFeatureEffectPythonAccuracy\% to
\SnapshotSolFeatureEffectPythonAccuracy\%. Recovering how features shape the
target, rather than what is wrong with the data table, is the main
weakness for Python-only agents. 

\begin{table}[t]
\caption{Python-only accuracy (\%) by model and template family.}
\label{tab:baseline}
\centering
\small
\begin{tabular}{lrrr}
\toprule
Model & Data quality & Feature contribution & Total \\
\midrule
\SnapshotBaselineRows
\midrule
\SnapshotBaselineAllRow
\bottomrule
\end{tabular}
\end{table}


\paragraph{The large search space of statistical discovery.}
Statistical discovery requires choosing not only which variables to examine
but also which analytical procedure to apply. For example, the same variables can be ranked in importance by many criteria, such as
marginal association, predictive contribution, or conditional effect, and
each criterion can produce a different ordering. An agent must therefore match its procedure
to the question's operational definition. The low feature-contribution accuracy
motivates examining how agents choose candidate relationships and
diagnostics, as illustrated by the case study in
Appendix~\ref{app:trajectories}.

\paragraph{Specifying an analysis from business language.}
Question wording changes the analytical specification available to the agent.
Business prompts can require translating an operational concern into a
statistical diagnostic, whereas data science prompts often name that diagnostic.
Python-only accuracy is \SnapshotDSPythonAccuracy\% for data science questions and drops to
\SnapshotBusinessPythonAccuracy\% for their business counterparts. 

\paragraph{Memorized knowledge versus evidence from the supplied table.}
Familiar column names and public-dataset regularities can suggest an answer before the agent even examines the edited table, but our interventions on the feature patterns make such priors insufficient for determining the reference answer.

\section{LLMs Augmented with Structured Data Analysis Tools}
\label{sec:results}

\subsection{Empirical Layer}

\begin{figure}[tbp]
\centering
\includegraphics[width=\linewidth]{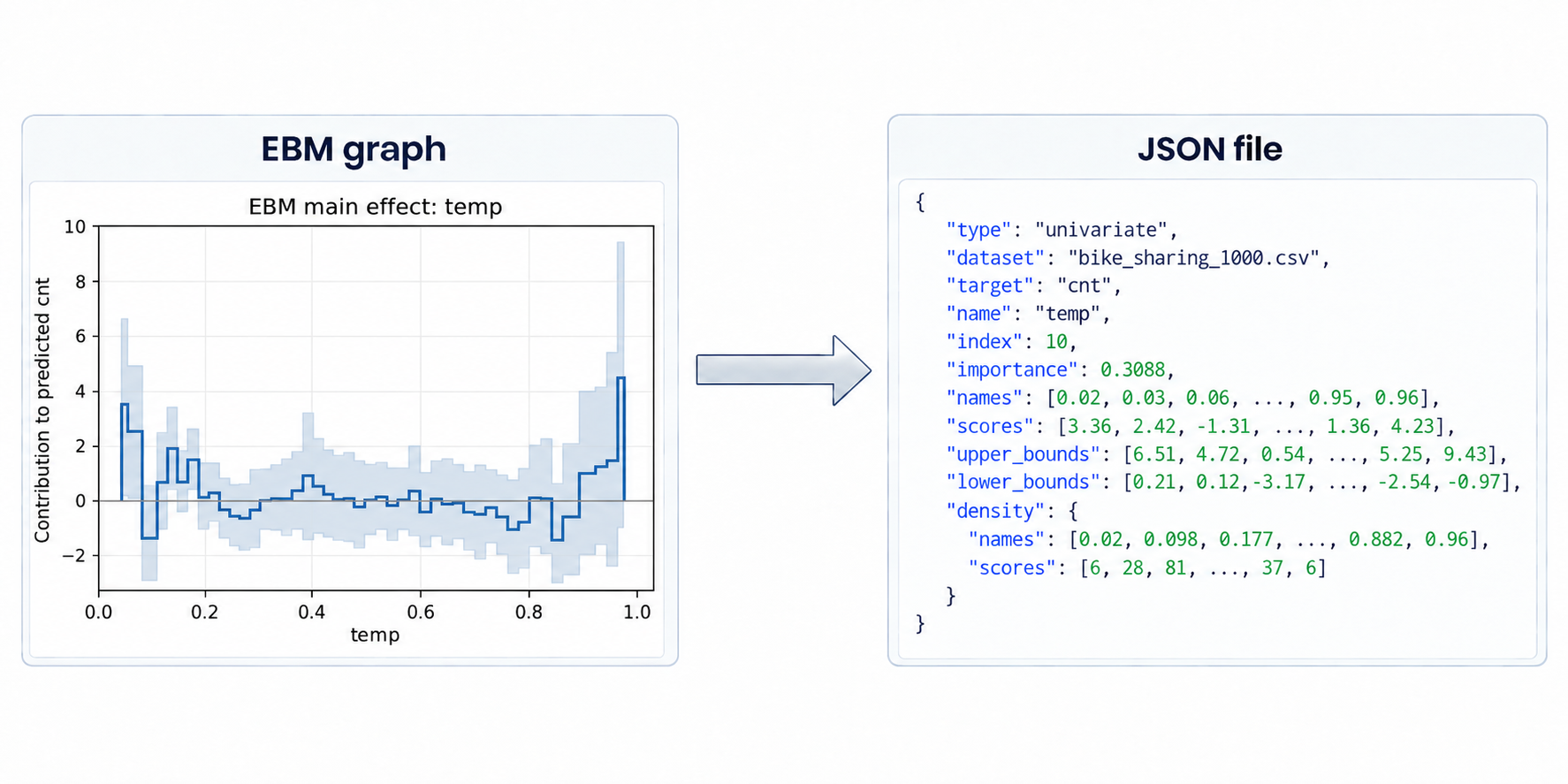}
\caption{Example of an Empirical Layer artifact from the Bike Sharing dataset.
The left panel shows the EBM single-feature effect of \texttt{temp} on
predicted bike rentals (\texttt{cnt}). The right panel is an excerpt from the JSON file storing this artifact.}
\label{fig:ebm-artifact}
\end{figure}

The previous section showed that, with Python alone, agents answer
data-quality questions accurately but struggle to recover feature contributions.
Motivated by this observation, we develop the Empirical Layer (Figure~\ref{fig:ebm-artifact}), which provides precomputed statistical summaries and learned representations of the supplied table. For each validated edited table,
we compute the same generic representation using only the table and its
designated target column. The
Empirical Layer comprises column statistics, distributions, correlations,
missing-value patterns, and outlier reports. Explainable Boosting Machines
(EBMs) additionally provide feature importance, fitted single-feature effects (shape functions),
and interaction effects \citep{nori2019interpretml}.


\subsection{Experimental Setup}

We compare \emph{Python alone} with \emph{Python + Empirical Layer} on the same question set in Section~\ref{sec:experiments}. The Empirical Layer tool is added alongside the Python tool; at each step, the agent chooses which tool to call. Throughout, $\Delta$ denotes
the P+E minus P difference in accuracy, in percentage points. Everything else, including the models, tables, questions, reference
answers, and grading, is identical to Section~\ref{sec:experiments}, so
every P response has a paired P+E response.

\subsection{Results}

\begin{figure}[t]
\centering
\includegraphics[width=\linewidth]{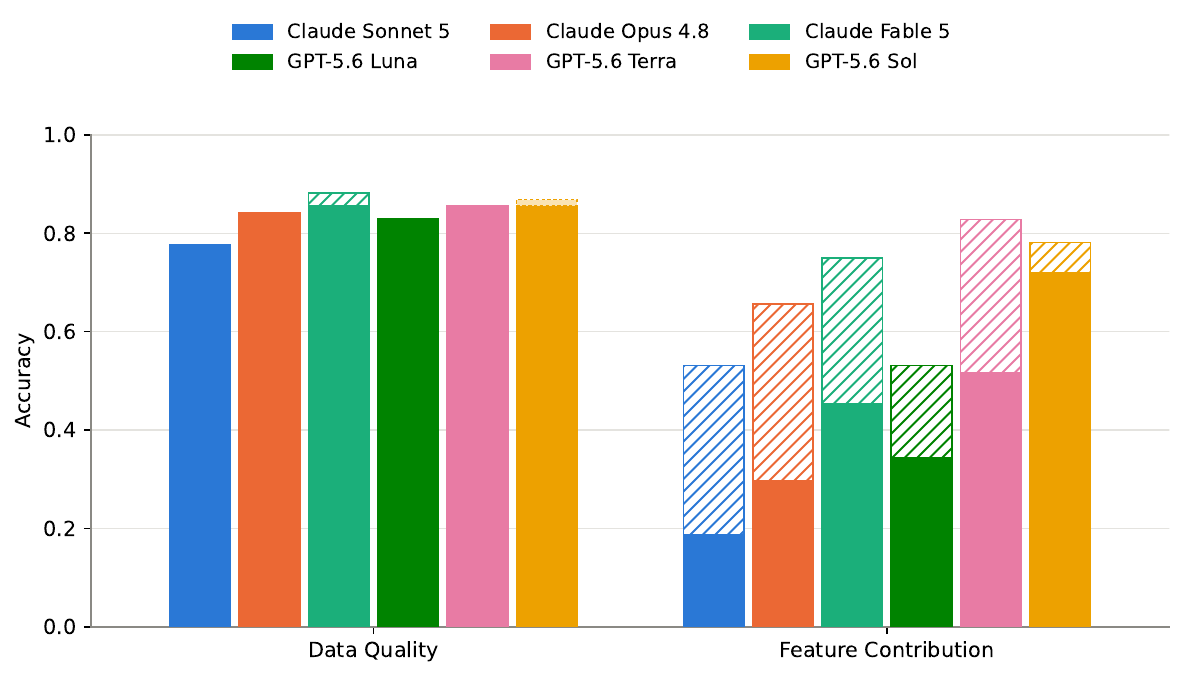}
\caption{Accuracy by template family and model. Solid bars show accuracy
with Python alone; hatched segments show the accuracy of Python +
Empirical Layer. Also see Table~\ref{tab:overall}
(Appendix~\ref{app:results}).}
\label{fig:family-model}
\end{figure}

\paragraph{Overall.}
Adding the Empirical Layer raises pooled accuracy from
\SnapshotPythonAccuracy\% to \SnapshotEmpiricalAccuracy\%, corresponding to
\SnapshotPythonCorrect{} versus \SnapshotEmpiricalCorrect{} correct answers (Figure~\ref{fig:family-model}; exact values in
Table~\ref{tab:overall}, Appendix~\ref{app:results}).

\paragraph{The gain comes from feature contributions.}
The improvement is concentrated in the family where Python-only agents
fail the most (Figure~\ref{fig:family-model}): Feature-contribution accuracy rises from
\SnapshotFeatureEffectPythonAccuracy\% to
\SnapshotFeatureEffectEmpiricalAccuracy\%
(\SnapshotFeatureEffectPythonCorrect{} versus
\SnapshotFeatureEffectEmpiricalCorrect{} correct answers). Data-quality accuracy is unchanged, moving
from \SnapshotDataQualityPythonAccuracy\% to
\SnapshotDataQualityEmpiricalAccuracy\% (a net gain of
\SnapshotDataQualityGain{} correct answer).

\paragraph{Differences across models.}
All six models improve, with overall gains ranging from \SnapshotSolDelta{}
to \SnapshotOpusDelta{} percentage points (Table~\ref{tab:overall}, Appendix~\ref{app:results}). On
feature-contribution questions the gains are larger, from
\SnapshotSolFeatureEffectDelta{} to \SnapshotOpusFeatureEffectDelta{}
percentage points, whereas on data-quality questions no model changes by
more than \SnapshotFableDataQualityDelta{} percentage points. The gain is
smallest for GPT-5.6 Sol, the strongest Python-only model, which already
reaches \SnapshotSolFeatureEffectPythonAccuracy\% on feature-contribution questions with
Python alone and moves to \SnapshotSolFeatureEffectEmpiricalAccuracy\%.

\begin{figure}[t]
\centering
\includegraphics[width=0.75\linewidth]{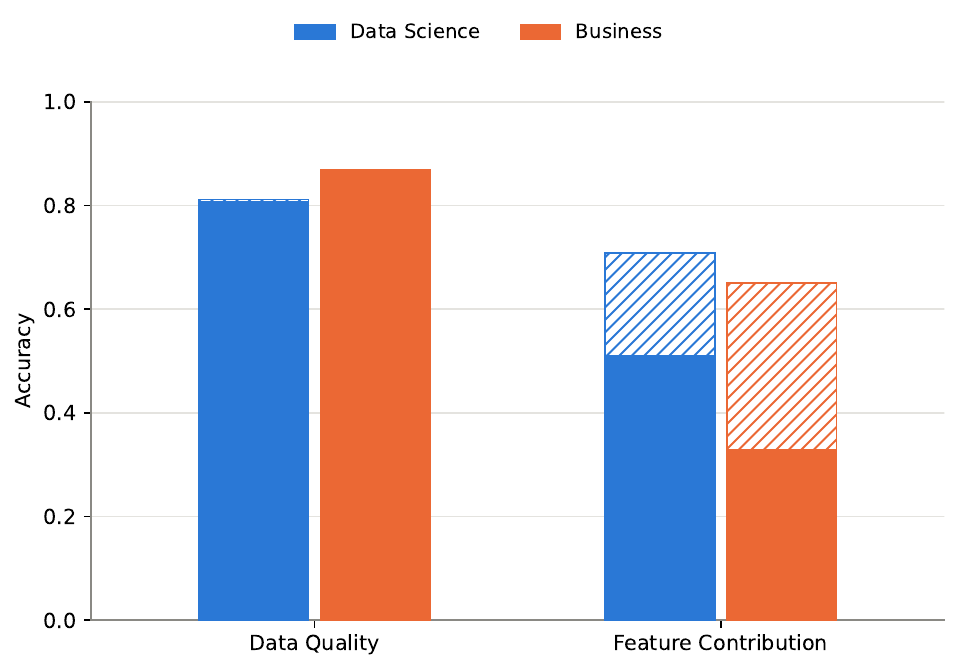}
\caption{Accuracy by template family and question framing, pooled over all
six models. Solid bars show accuracy with Python alone; hatched
segments show the accuracy of Python + Empirical Layer. Exact values
are listed in Tables~\ref{tab:framing} and~\ref{tab:family-framing}
(Appendix~\ref{app:results}).}
\label{fig:family-framing}
\end{figure}

\paragraph{Differences across question framings.}
Business questions gain more from the Empirical Layer than data science
questions: \SnapshotBusinessDelta{} versus \SnapshotDSDelta{} percentage
points overall (Figure~\ref{fig:family-framing}; exact values in
Table~\ref{tab:framing}, Appendix~\ref{app:results}). The difference comes
entirely from feature-contribution templates, where business questions gain
\SnapshotFeatureEffectBusinessDelta{} percentage points and data science
questions \SnapshotFeatureEffectDSDelta{}. With Python alone, agents do much worse on the business version of a
feature-contribution question than on its data science version
(\SnapshotFeatureEffectBusinessPythonAccuracy\% versus
\SnapshotFeatureEffectDSPythonAccuracy\%), and the Empirical Layer closes
most of that gap (\SnapshotFeatureEffectBusinessEmpiricalAccuracy\% versus
\SnapshotFeatureEffectDSEmpiricalAccuracy\%). A potential reason follows from
Section~\ref{sec:experiments}: a business prompt requires the agent to
translate an operational concern into a statistical diagnostic, so it must search a much larger space of statistical analyses and procedures. The Empirical Layer 
narrows that search significantly.


\section{Conclusion}
\label{sec:discussion}

Real-world data science requires agents not only to write code, but also to
discover which relationships in a table support an analytical conclusion. Public-table benchmarks can
also reward memorized answers, while fully synthetic tables may not retain the
background structure of real data. We introduced \sleuthbench to evaluate
statistical discovery through controlled, validated interventions. Injecting curated statistical evidence reduces the usefulness of memorized answers to the
original tables; reusable validators and automatic answer computation reduce manual
ground-truth curation while retaining source-table context. We also
introduced the Empirical Layer, which makes descriptive statistics, fitted EBM
effects, and feature metadata available as structured evidence for agents.
Across six LLMs, access to this layer raises feature-contribution accuracy from
\SnapshotFeatureEffectPythonAccuracy\% to
\SnapshotFeatureEffectEmpiricalAccuracy\% and pooled accuracy from
\SnapshotPythonAccuracy\% to \SnapshotEmpiricalAccuracy\%, while leaving
data-quality accuracy unchanged. Together,
\sleuthbench and the Empirical Layer provide a foundation for studying how
data-science agents find and use statistical evidence.

\clearpage

\subsection*{AI use statement}

In this work, we used generative AI tools to provide feedback on research methodology or experiments, interpret results, refine hypotheses, and to implement methods such as the grading process in the \sleuthbench pipeline. We have not used generative AI tools to assist with translation, clean and reformat datasets, or support qualitative and thematic data analysis. Generating synthetic data sets, helping develop theoretical models or conceptual frameworks, formulating mathematical claims, providing critical ingredients for proving mathematical claims, and assisting in the writing of proofs are not applicable to this work. Additionally, we used generative AI tools to create Figure~\ref{fig:ebm-artifact}, software code, draft parts of the research paper, edit the research paper to improve readability, identify relevant literature, format references, and suggest a structure for the research paper. We have reviewed all AI-assisted work. We take responsibility for the final content of this work, including text, claims, or artifacts produced with the aid of generative AI.

\bibliography{sleuthbench}
\bibliographystyle{iclr2027_conference}

\clearpage

\appendix
\section*{Appendix}
This appendix gives the detailed benchmark and evaluation specification.
Appendix~\ref{app:phenomena} catalogs the templates, injections, and paired
question wordings. Appendix~\ref{app:datasets} describes source data and
preprocessing; Appendix~\ref{app:validation} defines reference answers and
validation criteria; and Appendix~\ref{app:protocol} details the paired
evaluation and grading.
Appendix~\ref{app:results} reports detailed results.
Appendix~\ref{app:trajectories} presents a case study in which Empirical Layer artifacts help an agent identify the target interaction.

\section{Templates, Injections, and Questions}
\label{app:phenomena}

\sleuthbench defines 17 question templates across two families, implemented
by 14 injectors. A template is a reusable task specification that defines compatible
variables, an injection, two question framings, an answer format, and
validation criteria.

Each accepted instance pairs
data-science and business questions with a common table and a reference
answer. Table~\ref{tab:full-inventory} reproduces the data-science and
business question templates. Uppercase braced
placeholders are filled with instance-specific column names and values.

\begingroup
\footnotesize
\setlength{\tabcolsep}{4pt}
\renewcommand{\arraystretch}{1.0}
\begin{longtable}{@{}>{\raggedright\arraybackslash}p{0.11\linewidth}
                       >{\raggedright\arraybackslash}p{0.17\linewidth}
                       >{\raggedright\arraybackslash}p{0.31\linewidth}
                       >{\raggedright\arraybackslash}p{0.32\linewidth}@{}}
\caption{The 17 benchmark templates and their injections, with the data-science and
business question wording from the public template registry. Braced
placeholders are filled when each instance is constructed.}
\label{tab:full-inventory}\\
\toprule
Family & Template & What the injection changes & Question wording (data-science and business) \\
\midrule
\endfirsthead
\toprule
Family & Template & What the injection changes & Question wording (data-science and business) \\
\midrule
\endhead
\bottomrule
\endfoot
Data quality
& Bad-row indicator
& Corrupt a subset of target values and add a binary column whose positive
entries mark exactly those rows.
& \emph{Data-science:} You can apply exactly one rule based on a single column to exclude problematic rows before analyzing \texttt{\{OUTCOME\_COL\}}. Which column should the rule be based on? Return a single column name.\newline
\emph{Business:} Some records have already been marked for review in a yes-or-no column. Which column contains that review marker? Return a single column name. \\
& Conditional anomaly indicator
& Rewrite 5\% of outcomes so that they remain within the observed support but
become surprising given similar records. Add the true binary flag and
same-prevalence placebo flags.
& \emph{Data-science:} Among the binary 0/1 columns, one indicator has 1 exactly on rows whose \texttt{\{OUTCOME\_COL\}} outcome is conditionally inconsistent with otherwise similar records. The injected outcomes remain within the original target support. Which column is the indicator? Return a single column name.\newline
\emph{Business:} Among the 0/1 columns, one indicator's 1 values mark records whose \texttt{\{OUTCOME\_COL\}} outcome is inconsistent with otherwise similar records, even though the outcome itself remains within the dataset's observed range. Which column is that indicator? Return a single column name. \\
& Category target outlier
& Choose a sufficiently large category and shift its target values up or down
until its mean stands out from the other categories.
& \emph{Data-science:} Is there a category in \texttt{\{FEATURE\}} whose \texttt{\{OUTCOME\_COL\}} values are unusually high or low compared with the other categories? Return only the category label.\newline
\emph{Business:} Is there a category in \texttt{\{FEATURE\}} whose \texttt{\{OUTCOME\_COL\}} values stand out from the other categories? Return only the category label. \\
& Conditional bad rows
& Change the targets of a small set of rows so they remain plausible in the
overall target distribution but are unusual given their feature values.
& \emph{Data-science:} Some rows have \texttt{\{OUTCOME\_COL\}} values that are plausible globally but unusual given their feature values. Identify the \texttt{\{N\_INJECT\_SAMPLES\}} affected rows and return their row\_id values as a comma-separated list.\newline
\emph{Business:} There are \texttt{\{N\_INJECT\_SAMPLES\}} rows in this data with an unusual pattern in \texttt{\{OUTCOME\_COL\}}. Return the row\_id values as a comma-separated list. \\
& Underpowered group evidence
& Add a categorical group column with two categories: a common group A and a
very rare group B; leave the target unchanged so the group comparison has a
wide interval.
& \emph{Data-science:} For categorical feature \texttt{\{GROUP\_COL\}}, is there enough statistical evidence to distinguish group A from group B with respect to \texttt{\{OUTCOME\_COL\}}? Return one of: enough evidence, not enough evidence.\newline
\emph{Business:} For the categorical feature \texttt{\{GROUP\_COL\}}, is the evidence strong enough to say the groups behave differently with respect to \texttt{\{OUTCOME\_COL\}}? Return one of: enough evidence, not enough evidence. \\
& Semantic missing label
& Replace a small number of values in a categorical feature with a distinct
missing-like label such as ``Unknown'' or ``Not reported.''
& \emph{Data-science:} Which category label in \texttt{\{FEATURE\}} appears to encode missing, unknown, or not-reported values? Return only the label.\newline
\emph{Business:} Are there any labels in \texttt{\{FEATURE\}} that we need to pay attention to? If yes, return only the label. If no, return No. \\
& Target-dependent missing label
& Rename a categorical feature's categories, then place rows from one target tail into an additional neutral bucket. The moved rows lose their original category, so the bucket acts as an undocumented missing-value code. In real data, missing values are often recorded as ordinary-looking codes that the metadata does not identify, a problem known as disguised missing data~\citep{pearson2006disguised}.

& \emph{Data-science:} The category labels in \texttt{\{FEATURE\}} have been anonymised. One category label encodes missing or unknown values. Which missing-like label shows evidence of target-dependent missingness with respect to \texttt{\{OUTCOME\_COL\}}? Return only the label.\newline
\emph{Business:} Is there a category in \texttt{\{FEATURE\}} whose \texttt{\{OUTCOME\_COL\}} values stand out from the other categories? Return only the category label. \\
& Unreliable feature
& Add target noise whose variance increases with one numerical feature, making
the feature--target relationship much less consistent at one end.
& \emph{Data-science:} Which feature's relationship with \texttt{\{OUTCOME\_COL\}} is most unreliable --- i.e., has the highest prediction uncertainty? Return a single column name.\newline
\emph{Business:} For one column, \texttt{\{OUTCOME\_COL\}} results stay fairly close together at one end of the column but become much more spread out at the other end. Which column shows this uneven level of consistency? Return a single column name. \\
\midrule
Feature contribution
& Noise feature
& Permute one numerical feature across rows, preserving its marginal
distribution while breaking its row-wise relationship with the target.
& \emph{Data-science:} One numeric feature in this dataset has zero predictive importance for \texttt{\{OUTCOME\_COL\}} --- it contributes nothing to a predictive model. Which feature is it? Return a single column name.\newline
\emph{Business:} If we had to remove one number-based column before trying to estimate \texttt{\{OUTCOME\_COL\}}, which one could we remove with the least loss of useful information? Ignore columns used only to identify records. Return a single column name. \\
& Monotonicity reversal
& Add a V-shaped effect to a numerical feature whose original binned
relationship with the target was monotone, forcing a direction change.
& \emph{Data-science:} For each numeric feature, classify its relationship with \texttt{\{OUTCOME\_COL\}} as MONOTONE (consistently increasing or decreasing) or NON-MONOTONE (changes direction at least once). Which feature has the most non-monotone relationship with the outcome? Return only the feature name.\newline
\emph{Business:} For most number-based columns, \texttt{\{OUTCOME\_COL\}} generally keeps moving in one direction as the column value changes. Which column breaks that simple pattern most clearly? Return only the column name. \\
& Inverted-U peak
& Add a smooth effect centered at an interior feature value that declines on
both sides; the reference peak is measured on the edited table.
& \emph{Data-science:} One feature has an inverted-U relationship with \texttt{\{OUTCOME\_COL\}} --- performance peaks at a specific value and declines on both sides. Which feature, and at what value is the peak? Return your answer as: column\_name, value\newline
\emph{Business:} One column has a sweet spot for \texttt{\{OUTCOME\_COL\}}: results are best around a middle value and worse when the value is either lower or higher. Which column is it, and approximately where is the sweet spot? Return your answer as: column\_name, value \\
& Plateau threshold
& Add a clipped-linear effect: the target changes with a numerical feature up
to a threshold and then levels off.
& \emph{Data-science:} One feature has a threshold effect on \texttt{\{OUTCOME\_COL\}} --- it matters up to a point, then its effect levels off. Which feature, and at approximately what value does the effect plateau? Return your answer as: column\_name, value\newline
\emph{Business:} For one column, increasing its value is linked to changes in \texttt{\{OUTCOME\_COL\}} only up to a point. After that, further increases make little difference. Which column is it, and around what value does that happen? Return your answer as: column\_name, value \\
& Conditional direction
& Add an XOR-shaped target effect over median splits of two numerical features,
so their joint pattern dominates their marginal effects.
& \emph{Data-science:} Find the pair of features with the strongest interaction effect on \texttt{\{OUTCOME\_COL\}}. Sort them alphabetically. When the first feature is above its median, does increasing the second feature tend to increase or decrease \texttt{\{OUTCOME\_COL\}}? Return exactly: POSITIVE or NEGATIVE\newline
\emph{Business:} First find the two columns that reveal the strongest pattern when looked at together, and put their names in alphabetical order. Among records where the first column is in its upper half, does increasing the second usually make \texttt{\{OUTCOME\_COL\}} go up or down? Return exactly: POSITIVE for up or NEGATIVE for down. \\
& Dominant pair
& Add an XOR-shaped target effect over median splits of two numerical features,
so their joint pattern dominates their marginal effects.
& \emph{Data-science:} Which pair of features has the strongest interaction effect on \texttt{\{OUTCOME\_COL\}}? Return the two column names as: column\_a, column\_b\newline
\emph{Business:} Which two columns look fairly ordinary when considered separately but reveal the strongest unusual pattern in \texttt{\{OUTCOME\_COL\}} when looked at together? Return the two column names as: column\_a, column\_b \\
& Antagonistic reversal
& Add positive effects for each of two conditions separately but a negative
effect when both conditions hold.
& \emph{Data-science:} What are the two features whose combination has an effect opposite to each individual feature? Return the two column names as: column\_a, column\_b\newline
\emph{Business:} Which pairs of features behave unexpectedly when considered together? Return the two column names as: column\_a, column\_b \\
& Compensatory reversal
& Give the two conditions opposing individual effects, then add a positive
joint interaction that reverses one individual direction.
& \emph{Data-science:} What are the two features whose combination reverses the direction suggested by one individual feature? Return the two column names as: column\_a, column\_b\newline
\emph{Business:} Which pairs of features behave unexpectedly when considered together? Return the two column names as: column\_a, column\_b \\
& Positive synergy
& Give each of two conditions a positive individual effect and add a further
positive effect when both hold.
& \emph{Data-science:} What are the two features whose combination has an extra positive effect beyond each individual feature? Return the two column names as: column\_a, column\_b\newline
\emph{Business:} Which pairs of features behave unexpectedly when considered together? Return the two column names as: column\_a, column\_b \\
\end{longtable}
\endgroup

Across the six datasets, construction produces \SnapshotDatasetPairs{}
validated dataset--template combinations evaluated by all six models.
Template applicability and task-specific validation determine which
combinations enter evaluation.

\section{Source Data and Preparation}
\label{app:datasets}

The experiments use 1,000-row versions of six public datasets
(Table~\ref{tab:datasets}). Five sources are observational; AI4I is synthetic
data designed to reflect industrial predictive maintenance
\citep{ai4i2020dataset}. Together they provide transportation, housing,
maintenance, and energy-use settings with mixed numerical and categorical
variables.

\begin{table}[htbp]
\caption{Source datasets. Each experimental base table has 1,000 rows before
injection. Source row counts refer to the full prepared snapshots.}
\label{tab:datasets}
\centering
\small
\begin{tabularx}{\linewidth}{@{}>{\raggedright\arraybackslash}Xrl>{\raggedright\arraybackslash}p{0.20\linewidth}@{}}
\toprule
Dataset & Source rows & Target & Reference \\
\midrule
\href{https://archive.ics.uci.edu/dataset/601/ai4i+2020+predictive+maintenance+dataset}{AI4I 2020 Predictive Maintenance} & 10,000 & Machine failure & \citep{ai4i2020dataset} \\
\href{https://archive.ics.uci.edu/dataset/275/bike+sharing+dataset}{Bike Sharing} & 17,379 & Rental count & \citep{fanaee2013bike} \\
\href{https://scikit-learn.org/stable/modules/generated/sklearn.datasets.fetch_california_housing.html}{California Housing} & 20,640 & House value & \citep{sklearncalifornia} \\
\href{https://www.kaggle.com/datasets/harlfoxem/housesalesprediction}{King County Housing} & 21,613 & Sale price & \citep{kingcountyhousing} \\
\href{https://archive.ics.uci.edu/dataset/492/metro+interstate+traffic+volume}{Metro Interstate Traffic Volume} & 48,204 & Traffic volume & \citep{hogue2019metro} \\
\href{https://archive.ics.uci.edu/dataset/851/steel+industry+energy+consumption}{Steel Industry Energy Consumption} & 35,040 & Energy use & \citep{ve2021steel} \\
\bottomrule
\end{tabularx}
\end{table}

Metro and Steel timestamps are converted to month variables; Metro's
literal holiday category \texttt{None} becomes \texttt{No Holiday}.
The experiments use the 1,000-row versions of the standardized table.

Bike Sharing is associated with \citet{fanaee2014event}; California Housing
derives from the census data studied by \citet{pace1997sparse}.

\section{Construction and Validation Details}
\label{app:validation}

We construct each benchmark instance from a prepared base table and a
compatible question template. Construction has three steps: apply an
injection, validate the edited table, and compute the reference answer.

\paragraph{Template matching and injection.}
We match templates against the table's column summary. Each template
specifies eligible column types and cardinalities, dataset-level
requirements, question slots, and injector parameters. For a compatible
match, we assign columns and values to the slots and resolve the injector
parameters. We apply each distinct injector and parameter setting to a copy of the base table with a fixed random seed derived from the global seed.
The injector returns an edited table and a record of the realized
intervention. Each output table contains one intervention, although
multiple question templates can share it. We save the table with a
manifest recording the source, seed, target, parameters, intervention
record, slot assignments, and rendered question. The reference answer
remains unset. Incompatible templates and unsuccessful interventions
produce no instance.

\paragraph{Validation of the edited table.}
Validation is a separate stage that runs the validator registered for the injected phenomenon on each saved instance. The validator uses the intervention record
to identify the intended signal, then measures it on the edited table.
Acceptance criteria depend on the task. They can require a minimum number of rows in each compared group or bin, a specified shape or direction, sufficient signal strength, or
a margin over competing features or pairs when the question requires a
unique answer. The manifest
records each validation check's measurement, threshold, and decision, together with the overall validation result. Because a single random draw may not yield an unambiguous signal on a given table, some injectors try several candidate interventions, such as a different feature or a fresh random draw. They run the same validator on each candidate and keep the first one that passes, and they reject the injection if none passes within a fixed number of attempts. Since the validator reads only the table and the intervention record, the saved table is still validated independently before acceptance.

\paragraph{Reference-answer computation.}
For each validated instance, we call the answer function registered for
each applicable question template. The function receives the edited
table, the template's slot assignments, and the intervention record.
It writes the computed answer to the corresponding question entry in
the manifest. Answer functions are template-specific because different
questions can refer to the same intervention. Depending on the task,
the answer is a column, column pair, category, decision,
set of row identifiers, or column and value.

Reference answers follow explicit task definitions. Data-science and business phrasings
of the same template use the same intervention and answer definition.
Only questions with a passing validation result and a computed answer
enter model evaluation.

\section{Evaluation Protocol and Grading}
\label{app:protocol}

\subsection{Paired Evaluation and Coverage}

After the reference answers are computed, we evaluate six models: Claude Sonnet 5, Claude Opus 4.8, Claude Fable 5, and GPT-5.6 Sol, Terra, and Luna. Data-science and business questions are evaluated under two tool conditions: Python alone (P) and Python plus the Empirical Layer (P+E).

Each P/P+E pair consists of two responses to the same model–question instance, one under each tool condition. Data-science and business questions are treated as separate pairs.

Every model is evaluated on the same \SnapshotQuestionsPerModel{} questions:
data-science and business questions for \SnapshotDatasetPairs{}
dataset--template combinations. This gives \SnapshotResponses{}
model--question cases per condition and \SnapshotTotalRecords{} responses
in total. Data-science and business phrasings each contribute \SnapshotFramingResponses{}
cases per condition. 

The P+E condition permits retrieval of
structured statistics, EBM outputs, and language-based metadata. Agents
use the interface in \SnapshotEmpiricalUsage{}/\SnapshotResponses{} P+E
responses (\SnapshotEmpiricalUsagePercent\%), with
\SnapshotEmpiricalCalls{} retrieval calls in total. These comparisons
measure the effect of access to the combined interface.

\subsection{Grading}

Responses receive \textsc{Correct}, \textsc{Partial}, or
\textsc{Incorrect}; accuracy counts only \textsc{Correct}. Grades were assigned by a pipeline combining deterministic checks with
an LLM judge (\texttt{gpt-4o-mini}). Evaluation errors and empty answers were assigned
\textsc{Incorrect} without LLM judging. Unambiguous numerical and
column--value answers were checked directly against the reference answer;
remaining responses were assessed by the LLM judge using the question,
reference answer, and model response. 

Numerical values use a 5\% relative tolerance. A column--value answer must
also identify the correct column. \textsc{Partial} is used when a response
contains a correct component but does not fully answer the question. Examples
include an incomplete list or column pair, the correct qualitative direction
with an incorrect requested value, and the correct column paired with a value
outside the tolerance. A standalone numerical answer outside the tolerance is
\textsc{Incorrect}. A response that gives the correct answer remains \textsc{Correct} even if
it adds an explanation despite an answer-only instruction, provided the
explanation does not contradict or obscure the answer.

All reported results use the grades produced by this pipeline without manual
adjustment; the graded result files are preserved.

Execution failures are distinguished from missing evaluations. The
GPT runs that reached iteration or token limits have assigned \textsc{Incorrect} grades and
remain in the response counts.

\subsection{Prompt Formats}
\label{app:prompts}

We reproduce the system prompts verbatim. The evaluation prompt is given to
every evaluated model; the grading prompt is given to the LLM judge, followed
by the question, the reference answer, and the model response.

\begin{tcolorbox}[colback=sbblue, colframe=sbblue!30!black, title=Evaluation system prompt]
\small\itshape
``You are a data analyst. Answer the question about the dataset concisely.
Return ONLY the answer in the exact format requested with no extra text.
You may analyze only the preloaded pandas DataFrame \texttt{\textasciigrave df\textasciigrave}
and outputs returned by the declared tools. Do not inspect or access the
filesystem, directories, environment variables, network, subprocesses,
source code, manifests, metadata artifacts, or answer files. Do not use
open(), pathlib, os, glob, subprocess, socket, or shell commands.''
\end{tcolorbox}

\begin{tcolorbox}[colback=sbgray, colframe=sbgray!30!black, title=Grading system prompt]
\small\itshape
``You are grading an LLM's answer to a data analysis question about a CSV dataset.

\medskip
You will be given:\\
- The question asked\\
- The expected correct answer\\
- The model's answer

\medskip
Grade the response as exactly one of:\\
- CORRECT: The answer matches the expected answer (allow minor formatting
differences, extra explanation is fine as long as the core answer is right).
For a single numeric value, treat it as CORRECT when it is within 5\% relative
error of the expected value.\\
- PARTIAL: The answer is partially correct (e.g.\ some items in a list are
right, or the direction is right but the specific value is wrong)\\
- INCORRECT: The answer is wrong, irrelevant, or doesn't address the question

\medskip
Return JSON with \texttt{"grade"} and \texttt{"reasoning"} fields.''
\end{tcolorbox}

\FloatBarrier
\section{Detailed Results}
\label{app:results}

Each $N$ below counts model--question cases per condition; P and P + E are \textit{Python alone} and \textit{Python + Empirical Layer} as
in Section~\ref{sec:results}.

\begin{table}[htbp]
\caption{Accuracy (\%) by model and template family under P and P + E.}
\label{tab:overall}
\centering
\small
\setlength{\tabcolsep}{3.5pt}
\begin{tabular}{lrrrrrrrrr}
\toprule
& \multicolumn{3}{c}{Data quality} & \multicolumn{3}{c}{Feature contribution} & \multicolumn{3}{c}{Total} \\
\cmidrule(lr){2-4} \cmidrule(lr){5-7} \cmidrule(lr){8-10}
Model & P & P+E & $\Delta$ & P & P+E & $\Delta$ & P & P+E & $\Delta$ \\
\midrule
\SnapshotModelRows
\midrule
\SnapshotOverallRows
\bottomrule
\end{tabular}
\end{table}

\begin{table}[htbp]
\caption{Grade distributions by model and by family. C/P/I denotes
\textsc{Correct}/\textsc{Partial}/\textsc{Incorrect}.}
\label{tab:grade-distribution}
\centering
\small
\begin{tabularx}{\linewidth}{@{}Xrcc@{}}
\toprule
Model or family & $N$ & P: C/P/I & P+E: C/P/I \\
\midrule
\SnapshotGradeRows
\midrule
\SnapshotFamilyGradeRows
\bottomrule
\end{tabularx}
\end{table}

\begin{table}[htbp]
\caption{Accuracy by model and question framing. Every model is evaluated
on both framings of the same \SnapshotDatasetPairs{} dataset--template combinations.}
\label{tab:model-framing-results}
\centering
\small
\setlength{\tabcolsep}{4pt}
\begin{tabularx}{\linewidth}{@{}Xlrrrr@{}}
\toprule
Model & Framing & $N$ & P (\%) & P+E (\%) & $\Delta$ (pp) \\
\midrule
\SnapshotModelFramingRows
\bottomrule
\end{tabularx}
\end{table}

\begin{table}[htbp]
\caption{Accuracy by question framing, pooled over all six models and both
template families.}
\label{tab:framing}
\centering
\small
\begin{tabularx}{\linewidth}{@{}Xrrr@{}}
\toprule
Framing & P (\%) & P+E (\%) & $\Delta$ (pp) \\
\midrule
\SnapshotFramingRows
\bottomrule
\end{tabularx}
\end{table}

\begin{table}[htbp]
\caption{Accuracy by template family and question framing, pooled over all six
models.}
\label{tab:family-framing}
\centering
\small
\setlength{\tabcolsep}{4pt}
\begin{tabularx}{\linewidth}{@{}Xlrrrr@{}}
\toprule
Family & Framing & $N$ & P (\%) & P+E (\%) & $\Delta$ (pp) \\
\midrule
\SnapshotFamilyFramingRows
\bottomrule
\end{tabularx}
\end{table}

\section{Case Studies and Agent Trajectories}
\label{app:trajectories}

The Metro Interstate Traffic Volume case compares GPT-5.6 Luna's analyses
of the same dominant-interaction business question
(Table~\ref{tab:trajectory}). The reference pair is \texttt{month, temp}.
The injected effect is positive when exactly one feature exceeds its median
and negative otherwise, making the joint pattern central to the task.
Python-only exploration compares several interaction proxies, including
empirical mutual information, before selecting
\texttt{weather\_description, month}. With Empirical Layer access, Luna inspects
schema and metadata, retrieves target associations, and compares six EBM
interaction graphs before selecting the reference pair.

\begin{table}[htbp]
\caption{GPT-5.6 Luna on the Metro Interstate Traffic Volume business-language
dominant-interaction question. The reference pair is \texttt{month, temp}.}
\label{tab:trajectory}
\centering
\small
\begin{tabularx}{\linewidth}{@{}l>{\raggedright\arraybackslash}X>{\raggedright\arraybackslash}p{0.29\linewidth}l@{}}
\toprule
Condition & Analysis sequence & Final pair & Grade \\
\midrule
P & Compare interaction proxies, including empirical mutual information &
\texttt{weather\_description}, \texttt{month} & Incorrect \\
P+E & Inspect metadata and target associations; compare six EBM interaction graphs &
\texttt{month, temp} & Correct \\
\bottomrule
\end{tabularx}
\end{table}

The Python-selected pair has 28 weather categories and 12 month values,
allowing 336 joint cells for 1,000 rows. Sparse counts can favor such a pair
under uncorrected empirical mutual information, providing a possible
explanation for the final ranking. The P+E trajectory instead compares
fitted interaction terms under a common model. The \texttt{temp}--\texttt{month}
term has the largest interaction importance in the stored EBM artifacts,
consistent with the agent's answer.

\end{document}